# A MATHEMATICAL MODEL FOR LOGARITHMIC IMAGE PROCESSING

**Vasile PĂTRAȘCU, Vasile BUZULOIU**
Image Processing and Analysis Laboratory (LAPI)
"Politehnica" University Bucharest
E-mail: vpatrascu@tarom.ro, buzuloiu@alpha.imag.pub.ro

**ABSTRACT**
In this paper, we propose a new mathematical model for image processing. It is a logarithmical one. We consider the bounded interval (-1, 1) as the set of gray levels. Firstly, we define two operations: addition ⟨+⟩ and real scalar multiplication ⟨×⟩. With these operations, the set of gray levels becomes a real vector space. Then, defining the scalar product (.|.) and the norm ∥ . ∥, we obtain an Euclidean space of the gray levels. Secondly, we extend these operations and functions for color images. We finally show the effect of various simple operations on an image.

## 1. INTRODUCTION

An image is represented as a function defined on a bi-dimensional spatial domain. Images can be classified by the space in which these functions take their values in two groups: scalar images and vector images. Scalar images are defined as functions with real (bounded) values. The most frequent case in practice is when the value in a point (x,y) is the measure of the luminosity in that point (e.g. the TV images). Vector images are defined by real (bounded) vector functions. Nowadays color images are the most common particular case of vector images: R (red), G (green) and B (blue) components are considered vector real components. Scalar image values are called gray levels and the functions that define scalar images are called gray level functions. In order to use these gray levels as some algebra elements, the most frequently mathematical model used is the classical one, based on the real numbers algebra ( the case of linear processing [1], [2]). This leads to an implicit acceptance of the set of the gray level values as the whole real axis. The result is in contradiction with the fact that gray level functions are bounded. The practical solution to overcome this situation consists in truncating the values as soon as they go out of the true gray level set, be this truncation at the end of chain of operations or at each step. The problem is that in such a way we don't control in the right way the effect of the truncations. Within the mathematical model developed in this paper, the set of gray levels will be the bounded interval $(-1, 1)$ put in one-to-one correspondence with the physical interval of e.g. luminosities (0,M), by a linear transformation y=2(x-M/2)/M. The problem is to organize the set (-1,1) as a real vector space i.e. closed with respect to an addition (inner group operation) and to a scalar multiplication with real numbers. The key of this approach is to use an adequate isomorphism between R and (-1,1), respectively between R and (0,M) and the solution is of a logarithmic nature. There exist such earlier models. Stockham [7] proposed an image enhancement method based on the homomorphic theory introduced by Oppenheim [5] and applied it to images obtained by transmission or reflection of light through absorbing (reflecting) media, where the effects are naturally of multiplicative form. Another approach exists in the general setting of logarithmic representations suited for imaging processes by the transmitted light as well as for human perception: Jourlin and Pinoli introduced their logarithmic model in [4]. In both cases mentioned above the model works with semi-bounded sets for image values. This is perhaps the main reason which limits extending them to color image processing or to color spaces (especially to perceptual coordinates). We developed in [6] a logarithmic representation for gray levels images which works with bounded real sets and whose generalization for vector value images is straightforward. We shall develop it in the sequel of this paper and we will show the effect of various simple operations on an image. The paper is organized as follows: Section 2 and 3 introduces the addition and real scalar multiplication, the scalar product and the norm for the gray levels space respectively for the gray level images space. Section 4 and 5 introduces the addition and real scalar multiplication, the scalar product and the norm for the color space respectively for the color images space. Sections 6 and 7 presents some experimental results and Section 8 outlines the conclusions.

## 2. THE REAL VECTOR SPACE OF GRAY LEVELS

We consider as the space of gray levels, the set E=(-1,1). In the set of gray levels E we will define the addition ⟨+⟩ and the real scalar multiplication ⟨×⟩.

**Addition**
For $\forall v_1, v_2 \in E$ the sum $v_1 \langle + \rangle v_2$ is defined by the following relation:

$$v_1 \langle + \rangle v_2 = \frac{v_1 + v_2}{1 + v_1 \cdot v_2} \quad (1)$$

where the operations in the right side are meant in R.
The neutral element for addition is $\theta = 0$. Each element $v \in E$ has as its opposite the element $w = -v$ and this verifies the following equation: $v \langle + \rangle w = \theta$.

The addition $\langle+\rangle$ is stable, associative, commutative, has a neutral element and each element has an opposite. It means that this operation establishes on E a commutative group structure.

We can also define the subtraction operation $\langle-\rangle$ by:

$$v_1\langle-\rangle v_2 = \frac{v_1 - v_2}{1 - v_1 \cdot v_2} \quad (2)$$

Using subtraction $\langle-\rangle$, we will note the opposite of v, with $\langle-\rangle v$.

**Scalar multiplication**

For $\forall \lambda \in R, \forall v \in E$, we define the product between $\lambda$ and v by:

$$\lambda\langle\times\rangle v = \frac{(1+v)^\lambda - (1-v)^\lambda}{(1+v)^\lambda + (1-v)^\lambda} \quad (3)$$

where again the operations in the right hand side of the equality are meant in R.

The two operations, addition $\langle+\rangle$ and scalar multiplication $\langle\times\rangle$ establish on E a real vector space structure.

**The Euclidean space of the gray levels**

Further on, we will introduce the notions of scalar product and norm.

We define the scalar product $(.|.)_E : E \times E \to R$ by:

$$\forall v_1, v_2 \in E \quad (v_1 | v_2)_E = \varphi(v_1)\varphi(v_2) \quad (4)$$

where $\varphi : (-1,1) \to R$ and $\varphi(x) = \text{arcth}(x)$.

With the scalar product $(.|.)_E$ the gray levels space becomes an Euclidean space.

The norm $\|.\|_E : E \to [0, \infty)$ is defined via the scalar product: $\forall v \in E$, $\|v\|_E = ((v|v)_E)^{1/2} = |\varphi(v)| \quad (5)$

### 3. THE VECTOR SPACE OF THE GRAY LEVEL IMAGES

A gray level image is a function defined on a bi-dimensional compact D from $R^2$ taking the values in the gray level space E. We note with F(D,E) the set of gray level images defined on D. We can extend the operations and the functions from gray level space E to gray level images F(D,E) in a natural way:

**Addition**

$\forall f_1, f_2 \in F(D,E), \forall (x,y) \in D$,

$$(f_1\langle+\rangle f_2)(x,y) = f_1(x,y)\langle+\rangle f_2(x,y) \quad (6)$$

The neutral element is the identical null function.

The addition $\langle+\rangle$ is stable, associative, commutative, has a neutral element and each element has an opposite. As a conclusion, this operation establishes on the set F(D,E) a commutative group structure.

**Scalar multiplication**

$\forall \lambda \in R, \forall f \in F(D,E_3), \forall (x,y) \in D$,

$$(\lambda\langle\times\rangle f)(x,y) = \lambda\langle\times\rangle f(x,y) \quad (7)$$

The two operations, addition $\langle+\rangle$ and scalar multiplication $\langle\times\rangle$ establish on F(D,E) a real vector space structure.

**The Hilbert space of the gray level images**

Let f and g be two integrable functions from F(D,E). We define the scalar product by: for $\forall f_1, f_2 \in F(D,E)$

$$(f_1 | f_2)_{L^2(E)} = \int_D (f_1(x,y) \| f_2(x,y))_E \, dxdy \quad (8)$$

With the scalar product the gray level images space F(D,E) becomes a Hilbert space.

Further on, we will define the norm.

$$\forall f \in F(D,E), \|f\|_{L^2(E)} = \left(\int_D \|f(x,y)\|_E^2 \, dxdy\right)^{\frac{1}{2}} \quad (9)$$

### 4. THE REAL VECTOR SPACE OF THE COLORS

Next we consider as the space of colors, the cube $E_3(-1,1)^3$. We will note with r, g and b (red, green, blue) the three components of a vector $v \in E_3$. In the cube $E_3$ we will define the addition $\langle+\rangle$ and the real scalar multiplication $\langle\times\rangle$.

**Addition**

$\forall v_1, v_2 \in E_3$ with $v_1 = (r_1, g_1, b_1)$, $v_2 = (r_2, g_2, b_2)$, the sum $v_1\langle+\rangle v_2$ is defined by:

$$v_1\langle+\rangle v_2 = (r_1\langle+\rangle r_2, g_1\langle+\rangle g_2, b_1\langle+\rangle b_2) \quad (10)$$

The neutral element for addition is $\theta = (0,0,0)$.

Each element $v = (r,g,b) \in E_3$ has its opposite the element $w = (-r,-g,-b)$ and obviously $v\langle+\rangle w = \theta$.

The addition $\langle+\rangle$ is stable, associative, commutative, has a neutral element and each element has an opposite. It results that this operation establishes on $E_3$ a commutative group structure.

We can also define the subtraction operation $\langle-\rangle$ by:

$$v_1\langle-\rangle v_2 = (r_1\langle-\rangle r_2, g_1\langle-\rangle g_2, b_1\langle-\rangle b_2) \quad (11)$$

Using subtraction operation $\langle-\rangle$, we will note the opposite of v, with $\langle-\rangle v$.

**Scalar multiplication**

For $\forall \lambda \in R, \forall v = (r,g,b) \in E_3$, we define the product between $\lambda$ and v by:

$$\lambda\langle\times\rangle v = (\lambda\langle\times\rangle r, \lambda\langle\times\rangle g, \lambda\langle\times\rangle b) \quad (12)$$

The two operations, addition $\langle+\rangle$ and scalar multiplication $\langle\times\rangle$ establish on $E_3$ a real vector space structure.

**The Euclidean space of the colors**

Further on, we will introduce the notions of scalar product and norm.

We define the scalar product $(.|.)_{E_3} : E_3 \times E_3 \to R$ by:

$\forall v_1, v_2 \in E_3$ with $v_1 = (r_1, g_1, b_1)$, $v_2 = (r_2, g_2, b_2)$,

$$(v_1 | v_2)_{E_3} = \varphi(r_1) \cdot \varphi(r_2) + \varphi(g_1) \cdot \varphi(g_2) + \varphi(b_1) \cdot \varphi(b_2)$$

$$(13)$$

where $\varphi : (-1,1) \to R$ and $\varphi(x) = \text{arcth}(x)$.

With the scalar product $(.|.)_{E_3}$ the color space becomes a three-dimensional Euclidean space.

We define the norm $\|.\|_{E_3}: E_3 \to [0,\infty)$ by:

$$\forall v = (r,g,b) \in E_3, \|v\|_{E_3} = \sqrt{\varphi^2(r) + \varphi^2(g) + \varphi^2(b)} \quad (14)$$

## 5. THE VECTOR SPACE OF THE COLOR IMAGES

A color image is a function defined on a bi-dimensional compact D from $R^2$ taking the values in the colors space $E_3$. We note with $F(D,E_3)$ the set of color images defined on D. We extend the operations and the functions from colors space $E_3$ to color images $F(D,E_3)$ in a natural way:

**Addition**
For $\forall f_1, f_2 \in F(D,E_3)$, $\forall (x,y) \in D$,

$$(f_1 \langle + \rangle f_2)(x,y) = f_1(x,y) \langle + \rangle f_2(x,y) \quad (15)$$

The neutral element is the identical null function.
The addition $\langle + \rangle$ is stable, associative, commutative, has a neutral element and each element has an opposite. As a conclusion, these operations establish on the set $F(D,E_3)$ a commutative group structure.

**Scalar multiplication**
For $\forall \lambda \in R, \forall f \in F(D,E_3), \forall (x,y) \in D$,

$$(\lambda \langle \times \rangle f)(x,y) = \lambda \langle \times \rangle f(x,y) \quad (16)$$

The two operations, addition $\langle + \rangle$ and scalar multiplication $\langle \times \rangle$ establish on $F(D,E_3)$ a real vector space structure.

**The Hilbert space of the color images**
Let f and g be two integrable functions from $F(D,E_3)$. We define the scalar product by: $\forall f_1, f_2 \in F(D,E_3)$

$$(f_1 | f_2)_{L^2(E_3)} = \int_D (f_1(x,y) \| f_2(x,y))_{E_3} dxdy \quad (17)$$

With the scalar product defined, the color images space $F(D,E_3)$ becomes a Hilbert space.
Further on, we will define the norm.
$\forall f \in F(D,E_3)$,

$$\|f\|_{L^2(E_3)} = \left( \int_D \|f(x,y)\|_{E_3}^2 dxdy \right)^{\frac{1}{2}} \quad (18)$$

## 6. SIMPLE OPERATIONS USING THE MODEL

We used some images that are taken from [3],[8],[9]. Transformed images are shown with their equations of transformation. We can see in Fig. 1b and Fig. 2b that luminosity is changed by "addition" of a constant (in the sense of our model). For the image "lax", we modified the contrast by a scalar multiplication (Fig. 3b). For gray level image "Lena" we showed the negative $\langle - \rangle f$, positive part $f_+$ and negative part $f_-$ in Fig. 4b, 4c, 4d.
In Fig. 5 we show a correction for image "hats" using the subtraction operation. For color images "wildlife6" and "colors6" we modified the luminosity by addition of a constant and we modified the contrast and the saturation using the scalar multiplication ( see Fig.6b and 7b). In Fig. 8b we show the negative of color image "Lena" and in Fig.8c we obtained a color correction for the reddish image "Lena" where v=( 0.449, -0.241, -0.164). In Fig.9 we obtained a color correction for the reddish image "boat" where v=(-0.194, -0.323, -0.338). In Fig. 10 we exemplify the luminosity and color correction for the bluish image "puerta" where v=(-0.695, -0.495, -0.211). In Fig.11 we show the color correction for bluish image "gull" with v=(-0.865, -0.692, 0.210).

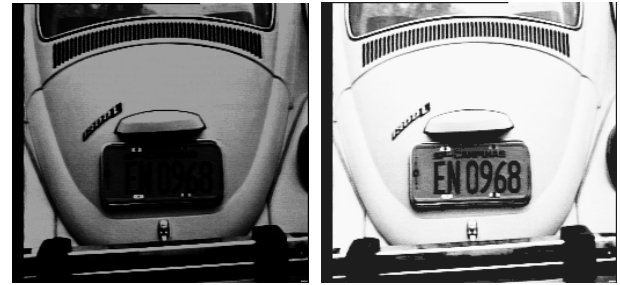

(a)          (b)

Fig.1    a) Image "fusca"
       b) Transformed with $t(f) = f \langle + \rangle 0.93$

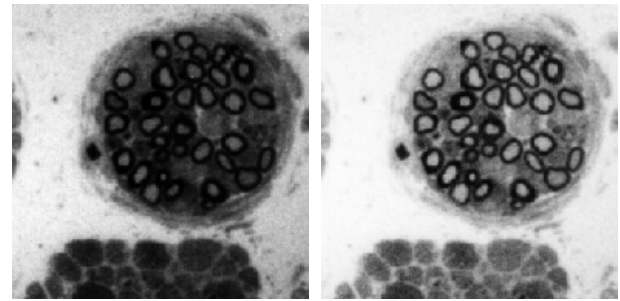

(a)          (b)

Fig.2    a) Image "myelin"
       b) Transformed with $t(f) = f \langle + \rangle 0.6$

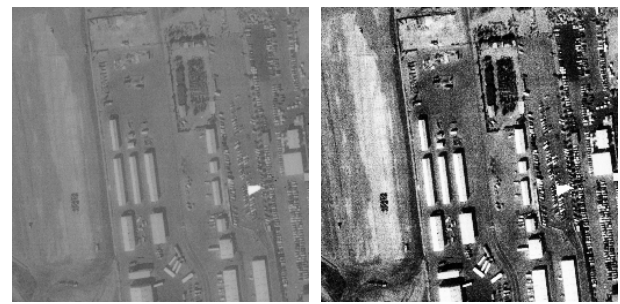

(a)          (b)

Fig.3    a) Image "lax"
       b) Transformed with $t(f) = 7 \langle \times \rangle f$

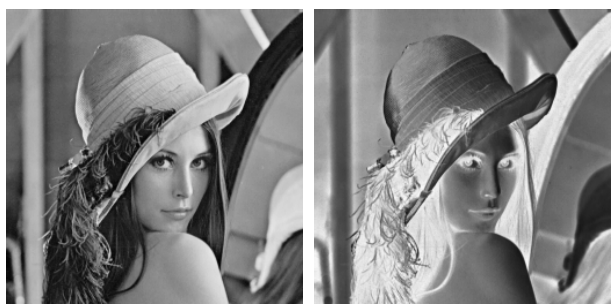

(a)      (b)

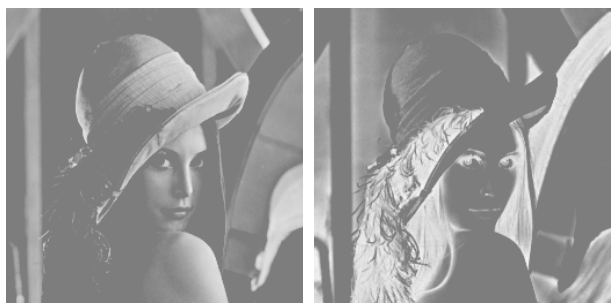

(c)      (d)

Fig.4   a) Image "Lena"
b) Transformed $t(f) = \langle - \rangle f$
c) Transformed $t(f) = f_+$
d) Transformed $t(f) = f_-$

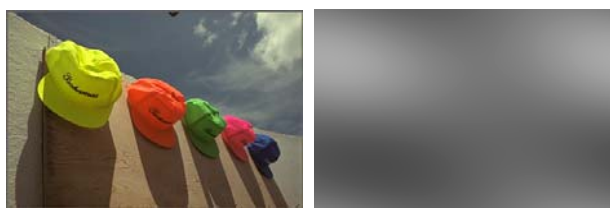

(a)      (b)

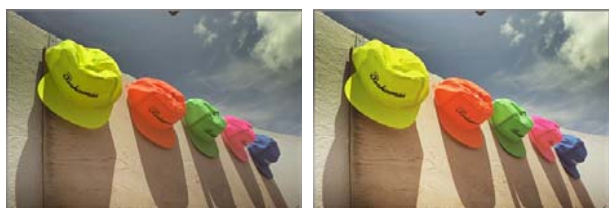

(c)      (d)

Fig.5   a) Image f ("hats")
b) Correction image $l_G$
c) Image $h = f \langle - \rangle l_G$
d) Transformed with $t(h) = 1.7 \langle \times \rangle h \langle + \rangle 0.2$

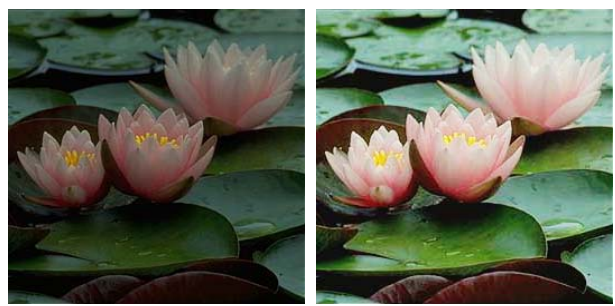

(a)      (b)

Fig.6   a) Image "wildlife6"
b) Transformed $t(f) = 2 \langle \times \rangle (f \langle + \rangle 0.5)$

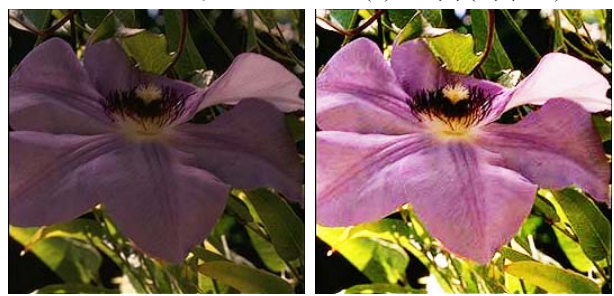

(a)      (b)

Fig.7   a) Image "colors6"
b) Transformed $t(f) = 3 \langle \times \rangle (f \langle + \rangle 0.5)$

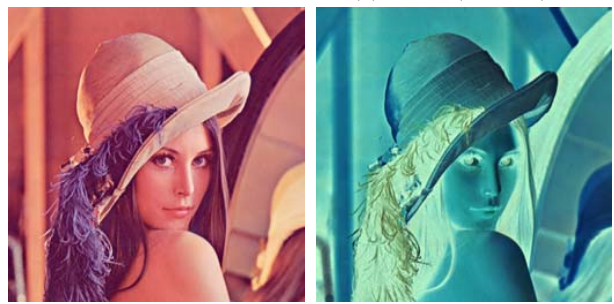

(a)      (b)

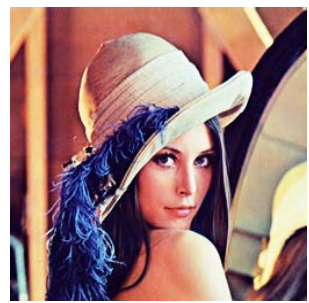

(c)

Fig.8   a) Image "Lena"
b) Transformed with $t(f) = \langle - \rangle f$
c) Transformed $g = 2.5 \langle \times \rangle (f \langle - \rangle 0.7 \langle \times \rangle v)$

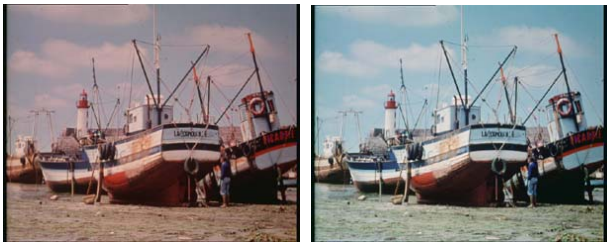

(a)     (b)

Fig.9  a) Image "boat"
b) Correction  t(f)=1.2⟨×⟩( f⟨-⟩1.5⟨×⟩v )

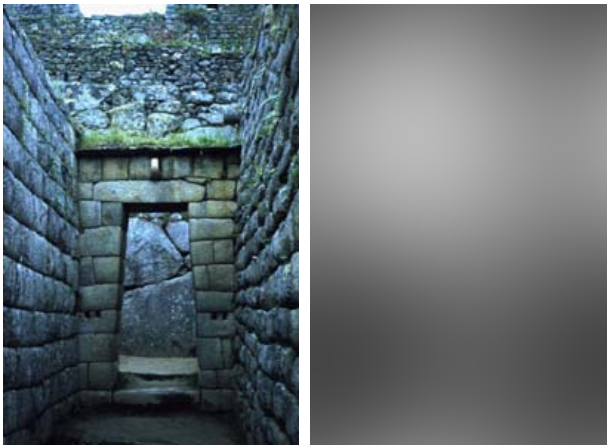

(a)     (b)

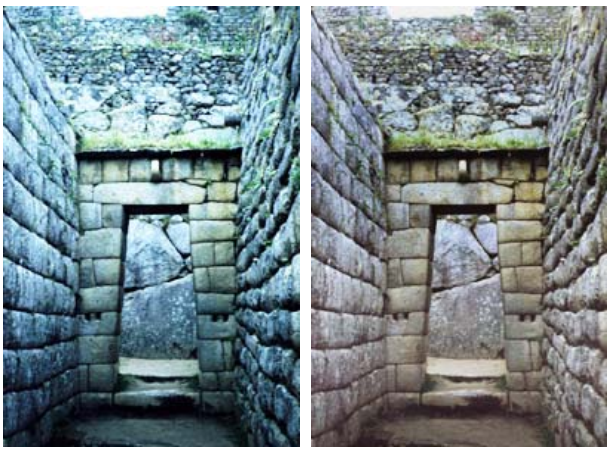

(c)     (d)

Fig.10 a) Image f  ("puerta")
b) Correction image $l_G$
c) Histogram equalization of "puerta"
d) Transformed  $t(f) = 1.43\langle\times\rangle f\langle-\rangle 1.39\langle\times\rangle v\langle-\rangle l_G$

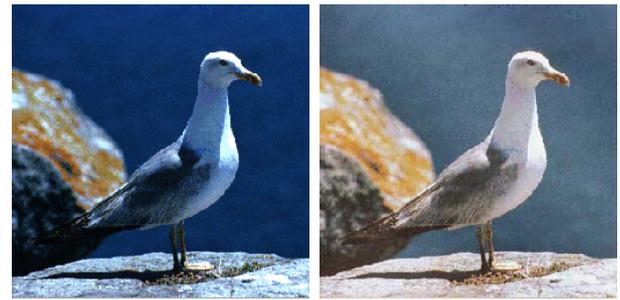

(a)     (b)

Fig.11  a) Image "gull"
b) Transformed  $t(f) = 0.95\langle\times\rangle f\langle-\rangle 0.76\langle\times\rangle v$

## 7. THE CONTRAST OF GRAY LEVEL IMAGES

The structure of vector space defined for set F(D,E) allows us to translate the notion of contrast from classical framework. We will note $p_i=(x_i,y_i)$, the pairs of coordinates that define the spatial position of a pixel in an image.

**The relative contrast between two pixels**

The relative contrast between two distinct pixels $p_1, p_2 \in D$, for an image, $f : D \to E$ is a logarithmic gray levels image noted $C_R(p_1, p_2)$, defined by relation:

$\forall p_1, p_2 \in D, p_1 \neq p_2$,

$$C_R(p_1, p_2) = \frac{1}{d(p_1, p_2)} \langle \times \rangle \frac{f(p_1) - f(p_2)}{1 - \frac{f(p_1) \cdot f(p_2)}{M^2}} \quad (19)$$

where $d(p_1,p_2)$ is the Euclidean distance between $p_1$ and $p_2$ in the $R^2$ plain.

**The absolute contrast between two pixels**

From the relative contrast $C_R$ we can defined the absolute contrast by the formulas:

$\forall p_1, p_2 \in D, p_1 \neq p_2$,

$$C_A(p_1, p_2) = |C_R(p_1, p_2)| = \frac{1}{d(p_1, p_2)} \langle \times \rangle \frac{|f(p_1) - f(p_2)|}{1 - \frac{f(p_1) \cdot f(p_2)}{M^2}}$$
(20)

**The contrast for a pixel**

The contrast for an arbitrary pixel $p \in D$, for an image $f \in F(D, E)$ is a positive logarithmic gray levels image, noted C(p), defined by the mean of absolute contrast between the pixel p and the pixels $(p_i)_{i=1,n}$ that belong to a neighborhood V. Thus we have the next formula:

$$\forall p \in D, \qquad C(p) = \frac{1}{n} \langle \times \rangle \left( \underset{i=1}{\overset{n}{\langle + \rangle}} C_A(p, p_i) \right) \quad (21)$$

**Experimental results for the contrast**

In the next figures we show the contouring of images with the contrast formulae defined above. We define C(p)

as the contour image for an initial image, which obtained for each pixel using the contrast formulae.

In Fig. 12b, 12c, 12d and 13b, 13c, 13d we represented the horizontal relative contrast $C_R(p_{i,j+1}, p_{i,j})$, the vertical relative contrast $C_R(p_{i,j}, p_{i+1,j})$ and the contrast $C(p_{i,j})$.

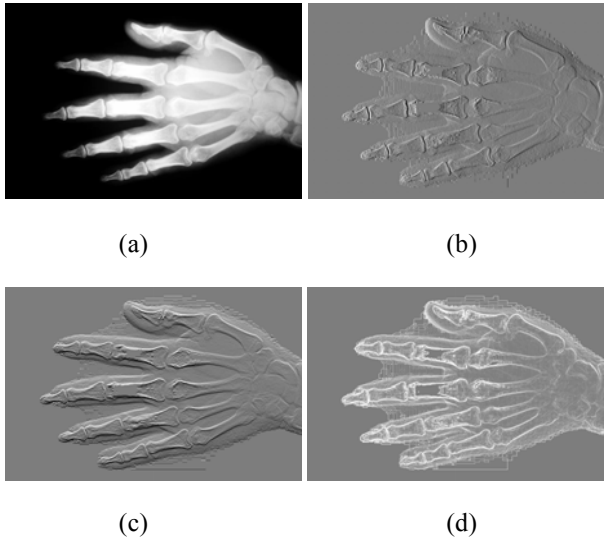

(a) (b)

(c) (d)

Fig.12

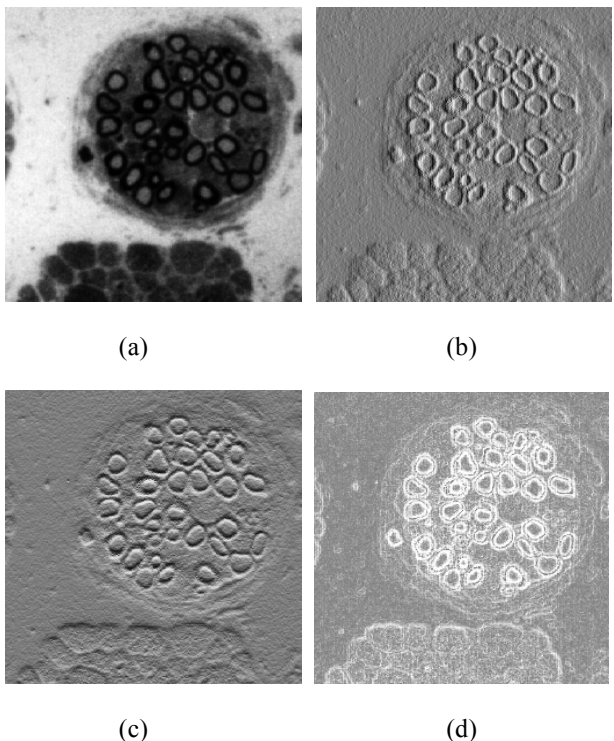

(a) (b)

(c) (d)

Fig. 13

## 8. CONCLUSIONS

We have presented in this paper a new mathematical model for gray level images and also for color images. The main idea is to define an algebraical structure on a bounded interval.

The examples offered above show rather clearly that the operations (addition, scalar multiplication) and the functions (scalar product, norm) ground an important mathematic model for image processing.

Using this mathematical model we can obtain very simple algorithms for images processing.

## 9. REFERENCES

[ 1] K.R. Castleman. *Digital Image Processing*. Prentice Hall, Englewood Cliffs NJ 1996.
[ 2] A.K. Jain. *Fundamentals of Digital Image Processing*. Prentice Hall Intl., Englewood Cliffs NJ , 1989.
[ 3] R. Jordan, R. Lotufo, *Khoral Research, ISTEC*.
[ 4] M. Jourlin, J.C. Pinoli. *Modelisation & traitement des images logarithmiques*. Publication Nr. 6, Departement de Mathematiques, Univ. de Saint-Etienne 1992.
[ 5] A.V. Oppenheim. *Generalized superposition*. Information and control, 11, 528-536, 1967.
[ 6] V. Pătrașcu. *A mathematical model for logarithmic image processing*. PhD thesis, "Politehnica" University Bucharest, May 2001.
[ 7] T.G. Stockham. *Image processing in the context of visual model*. Proc. IEEE , Vol. 60, July 1972.
[ 8] *http://dragon.larc.nasa.gov/retinex.*
[ 9] *http://www.cse.usf.edu.*